\documentclass{svproc}
\usepackage{url}
\usepackage{epsfig}

\begin{document}
\mainmatter              
\title{Cross-spectral Periocular Recognition: A Survey}
%
%
\author{S. S. Behera\inst{1}, Bappaditya Mandal\inst{2}
\and N. B. Puhan\inst{1}}
\authorrunning{Behera \emph{et al.}} 
%
%
\institute{School of Electrical Sciences, Indian Institute of Technology, Bhubaneswar, India.\\
\email{\{ssb11,nbpuhan\}@iitbbs.ac.in},
\and
School of Computing and Mathematics, Keele University, United Kingdom.\\
\email{b.mandal@keele.ac.uk}
}

\maketitle              

\begin{abstract}

Among many biometrics such as face, iris, fingerprint and others, periocular region has the advantages over other biometrics because it is non-intrusive and serves as a balance between iris or eye region (very stringent, small area) and the whole face region (very relaxed large area). Research have shown that this is the region which does not get affected much because of various poses, aging, expression, facial changes and other artifacts, which otherwise would change to a large variation. This region can be captured using the similar setups used for obtaining face and iris images. Active research has been carried out on this topic since past few years due to its obvious advantages over face and iris biometrics in unconstrained and uncooperative scenarios. Many researchers have explored periocular biometrics involving both visible (VIS) and infra-red (IR) spectrum images. For a system to work for 24/7 (such as in surveillance scenarios), the registration process may depend on the day time VIS periocular images (or any mug shot image) and the testing or recognition process may occur in the night time involving only IR periocular images. This gives rise to a challenging research problem called the cross-spectral matching of images where VIS images are used for registration or as gallery images and IR images are used for testing or recognition process and vice versa. After intensive research of more than two decades on face and iris biometrics in cross-spectral domain, a number of researchers have now focused their work on matching heterogeneous (cross-spectral) periocular images. Though a number of surveys have been made on existing periocular biometric research, no study has been done on its cross-spectral aspect. This paper analyses and reviews current state-of-the-art techniques in cross-spectral periocular recognition including various methodologies, databases, their protocols and current-state-of-the-art recognition performances.
\end{abstract}
\section{Introduction}
Periocular region is a subset of the facial region which includes eyes as well as the area in the vicinity of eyes. This includes regions such as eyes, eye lashes, eyebrows and cheek or skin texture. Research work on person identification using periocular region got started ever since Park \textit{et al.} investigated its benefits over face and iris biometrics in \cite{Park-2009} in the year 2009.

Person recognition using face requires the full face to be available at the time of recognition. However, this never happens in practical surveillance scenarios due to the occlusion by hairs, facial accessories and so on. Not only occlusion but face biometrics also suffers from other major factors such as variation in pose, expression, age, facial changes and improper illumination \cite{Mandal2}, \cite{Mandal3}, \cite{Mandal4}. Performance of face recognition degrades in presence of such artifacts and has been haunting researchers from diverse areas of research for many decades \cite{mandal2016trends}. \cite{url:1} show cases how a 10 year old kid's face unlocks his mother's iPhone X in a few minutes time as his first attempt and justifies the fragility of face recognition technology and also described here for other interesting applications \cite{Mandal5}. Periocular biometrics alleviates the challenges associated with face recognition. Not only face but iris recognition also suffers from limitations such as requirement of images of very high resolution and high user cooperation in order to perform well \cite{Bharadwaj-2010}. Whereas, periocular recognition is less affected by such problems. Also, for iris recognition, IR images are preferred over VIS images because of less changes by uneven illumination, but, periocular biometrics performs well both in presence of IR and VIS light images.    

Cross-spectral biometrics refers to matching images of a person in two different spectra. This means that during registration, only VIS (day time) images are used and during testing or recognition, IR (night time) images are used and vice versa. This finds enormous use in the field of advanced surveillance applications where the query images are captured in a wavelength other than that of visible light.  A number of research works have been done on matching IR images with VIS images and vice-versa. The IR images include images from near IR (NIR), short-wave IR (SWIR), middle-wave IR (MWIR), and long-wave IR (LWIR) range. 

\begin{figure}[h!]
\centering
\includegraphics[width=1.2\linewidth]{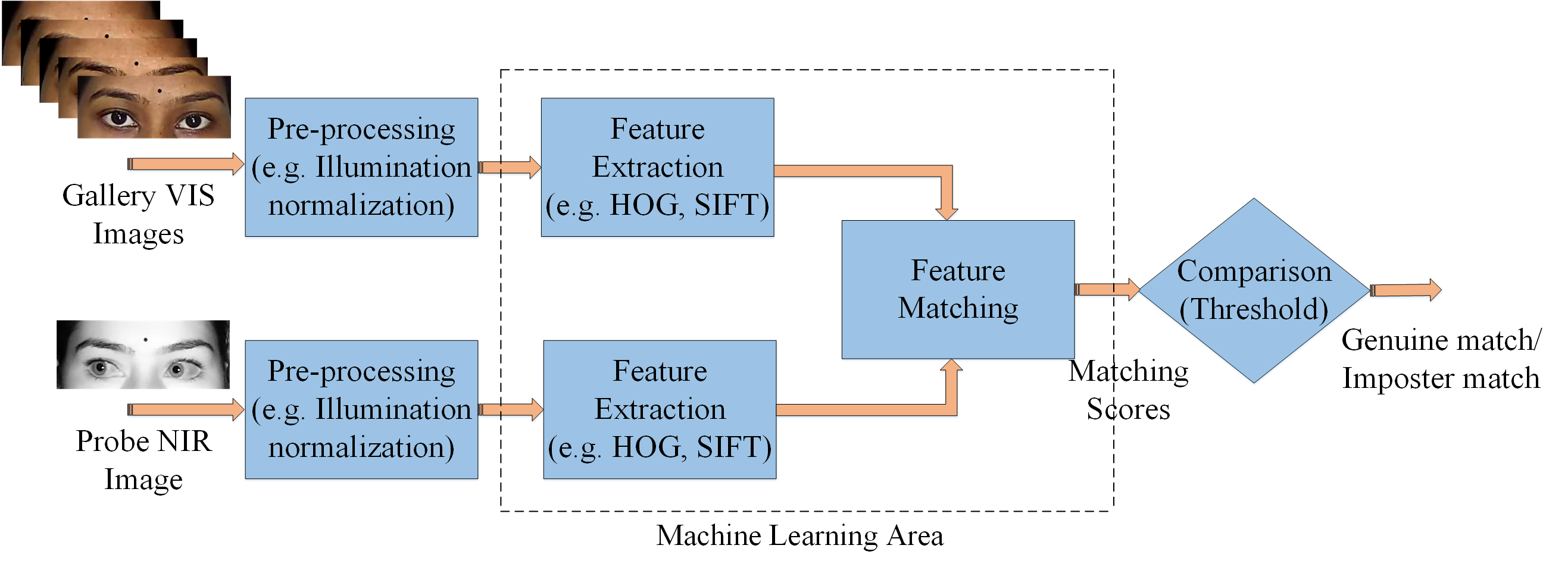}
\caption {A general block diagram showing the steps involved in cross-spectral periocular recognition. Two examples of commonly used features in periocular recognition are histogram of oriented gradients (HOG) based on the shape of edges and scale invariant feature transform (SIFT) feature based on key points in images. The dotted region shows the general region where researchers would incorporate machine learning algorithms.}
\label{fig:blockdiag}
\end{figure}

Fig. \ref{fig:blockdiag} shows the general block diagram of a recognition system used for cross-spectral matching of periocular images. We have shown a case where the gallery and probe consist of VIS images and an  NIR image respectively. As can be seen from the block diagram, the first step involved is called pre-processing where the images are processed in order to mitigate the effect of illumination and appearance variation present between them. One of the common pre-processing step includes illumination normalization. The prepossessed images are then fed to the feature extraction block in order to extract local or global descriptors. The gallery and probe images are then compared against each other using distance metrics in order to get normalized matching scores in a range of 0 to 1 where 0 means no match at all and 1 means a full match. The features can also be fed to a classifier (e.g. support vector machine or SVM) in order to get the probability of a probe image belonging to a particular class. Either the matching scores or the probability scores can be compared with a threshold to finally determine whether the probe is a genuine user or an imposter.

The performance of a biometric recognition system can be calculated in terms of genuine acceptance rate (GAR), defined as the rate at which a system accepts a genuine user as genuine, false rejection rate (FRR), defined as the rate at which the system rejects a genuine user considering it as imposter and false acceptance rate (FAR), defined as the rate at which the system accepts an imposter as genuine. Mathematically,
\begin{equation}
    GAR = \frac{Ture Positive}{True Positive + False Negative}
\end{equation}
\begin{equation}
    FRR = \frac{False Negative}{True Positive + False Negative}
\end{equation}
\begin{equation}
    FAR = \frac{False Positive}{False Positive + True Negative}
\end{equation}

Higher FRR values let more genuine users rejected by the system whereas higher FAR values let more imposter users access the system. Hence, it is required that both error rates (FRR and FAR) be be as small as possible, with FAR values must be smaller as compared to FRR values, in order to make the system more secure.

In the existing literature, a number of surveys are available in the context of periocular biometrics recognition \cite{Alonso-2016}, \cite{Nigam-2015}. but there has not been any survey done on cross-spectral periocular recognition. In this paper, we are the first to present a survey on cross-spectral periocular recognition. The existing methods and algorithms are presented in Section 2. Section 3 gives a detailed description of the existing databases and Section 4 presents the conclusion and future work.

\section{Methodologies Used in Cross-spectral Periocular Recognition}

The problem of cross-spectral matching has been studied in the context of face \cite{Hyunju-2011}, \cite{zhu-2014}, \cite{Juefei-2015} and iris \cite{Ross-2009}, \cite{Zuo-2010}, \cite{Ramaiah-2017} biometrics. So far a few works have studied this problem in the context of periocular region. Research on cross-spectral periocular recognition got attention ever since both Sharma \textit{et al.} \cite{Sharma-2014} and Cao and Schmid \cite{cao:1} contributed their works in the year 2014.

Authors in \cite{Sharma-2014} have proposed a method based on neural network based learning in order to match the images belonging to VIS, NIR, and night vision spectra. They have used two neural networks where each network learns weights for a particular spectrum. The two learned networks are then combined and trained on the cross-spectral images. Inputs to the neural networks are pyramid of histogram of oriented gradients (PHOG) features extracted from the images. A number of experiments have been performed on their home IIIT Delhi multi-spectral periocular (IMP) database which is the first database that comprises multi-spectral periocular images. In addition to that, the authors have compared their proposed approach with the experimental results obtained from matching images using four existing descriptors, where the PHOG descriptors outperforms the existing descriptors. When both left and right periocular images are taken for recognition, this method achieves GARs of $47.08$\%, $71.93$\%, and $48.21$\% at $1$\% FAR for VIS-NIR, VIS-night vision and NIR-night vision matching respectively.  

Cao and Schmid in \cite{cao:1} have proposed a method to evaluate cross-spectral periocular recognition based on Gabor filtering followed by extraction of local binary pattern (LBP) and generalized LBP (GLBP) features. The authors have performed their experiments on three different face datasets comprising NIR, SWIR, and MWIR images and at various standoff distances. The first dataset, tactical imager for night/day extended-range surveillance (TINDERS), is composed of VIS, NIR and SWIR images at two standoff distances ($50 m$ and $106 m$). The pre-TINDERS dataset is composed of images from VIS and SWIR wavelength acquired at a standoff distance of $1.5 m$. The third dataset, collected by Pinellas Country Sheriff's office (OPSC), is composed of color and MWIR images at $1.5 m$ distance. The authors have also compared the results of their proposed approach with two other algorithms known as LBP and Gabor ordinal measures (GOM). The highest reported accuracies obtained in terms of GAR values in matching SWIR-VIS, NIR-VIS and MWIR-VIS periocular images are 0.75 at an FAR of 0.01, 0.44 at an FAR of 0.1 and 0.35 at an FAR of 0.1 respectively.

Authors in \cite{cao:1} have extended their work in \cite{cao:2}, and \cite{cao:3}. In \cite{cao:2}, a study involving partial face at short and long standoff distances is presented. The whole face region is divided into three non-overlapping regions namely, (1) eyes and nasal bridge, (2) nasal tip and cheeks and (3) mouth and chin and the performance of each of these parts is evaluated. The additional experiments involving partial face include covering two out of three parts of the face 
and sequential covering of partial faces. 
The authors have performed a number of experiments considering different standoff distances as well as different spectra for each of these experiments. However, they have not presented the results in a tabular form in order to make the reader understand the purpose of their experiments. Their results show that experiments involving eyes and nasal bridge region achieve higher accuracies as compared to other two regions. Also the sequential covering experiments show that the full face outperforms the partial face regions.

In \cite{cao:3}, a compound operator, termed as GWLH, that uses Gabor filtering followed by fusion of three descriptors namely, Weber local descriptor (WLD), LBP, and HOG is introduced. The authors have performed both verification and identification experiments as well as considered the matching of VIS and LWIR images. The GAR values obtained for verification experiments are $93.88$\%, $98.05$\%, $42.39$\% and $29.44$\% at an FAR of 0.1 for matching VIS gallery images with SWIR, NIR, MWIR and LWIR images respectively. Similarly, the highest rank-1 identification rates obtained for matching VIS gallery images with SWIR, NIR, MWIR and LWIR images are $68.75$\%, $70.31$\%, $5.58$\% and $8.09$\% respectively. The proposed GWLH operator outperforms other existing operators except for the case of identification experiments for matching VIS-LWIR images, where it achieves an accuracy of $5.78$\%.

Ramaiah and Kumar \cite{Ramaiah-2016} have proposed a method based on Markov random fields (MRFs) and three patch LBP (TPLBP) descriptor. The authors have used exact pixel correspondences during image acquisition and synthesized NIR image pixels from the corresponding VIS periocular images. Two variants of LBP descriptors namely, three patch LBP (TPLBP) and four patch LBP (FLBP) are then extracted from the images. The method has been evaluated on IMP and PolyU dataset using both descriptors. TPLBP descriptor outperforms the FLBP operator by achieving a GAR of $73.2$\% and $18.35$\% at $1$\% FAR on PolyU and IMP database respectively.

Raja \textit{et al.} \cite{Raja-2016} have proposed a method to match NIR (probe) and VIS (gallery) periocular images using bank of statistical filters. They have first extracted binarized statistical image features from both gallery and probe images, and then compared them by calculating $\chi^2$ distance between them. The comparison scores from all filters are then fused in order to get the final score. The experiments are performed on their own cross-spectrum database consisting of 30 subjects where each subject consists of 8 sample images per eye instance per spectrum. Their proposed scheme achieves a genuine match rate (GMR) of $96.40$\% at an false match rate (FMR) of $0.01$\%.

While the research in periocular biometrics had taken root in cross-spectral scenario, Ana \textit{et al.} \cite{Crosseyed:1} made it stronger by conducting a competition called the first cross-spectrum iris or periocular recognition competition in 2016. The paper also describes development of a new cross-spectral periocular database consisting of both VIS and NIR images. An evaluation metric ($GF2$) involving generalized false accept rate (GFAR) at generalized false reject rate (GFRR) of $1$\% is used in order to evaluate the performance of the submitted algorithms. Two participants namely, \textbf{HH} from Halmstad University, Sweden and \textbf{NTNU} from Norwegian Biometrics Laboratory, Norwegian University of Science and Technology, Norway submitted their algorithms where one of the algorithms from HH, called $HH_1$ obtained a $GF2$ error of $0.00$\% and thus achieved nearly $100$\% accuracy in periocular verification. 

Inspired by the success of this work, the competition has been extended in the year 2017 using an enlarged version of the Cross-eyed database \cite{Crosseyed:2}. This paper includes 12 algorithms submitted by five teams namely, \textbf{HH} from Halmstad University, \textbf{NTNU} from Norwegian Biometrics Laboratory, \textbf{IDIAP} from Switzerland, \textbf{IIT Indore} from India and \textbf{Anonymous}. Among the algorithms submitted, $HH_1$ method achieved the best accuracy by obtaining a $GF2$ value of $0.74$\%. 

Raja \textit{et al.} \cite{Raja-2017} have addressed the problem of cross-spectral periocular verification using steerable pyramid features and a multi-class SVM classifier. The authors have further performed fusion of decisions and scores obtained from steerable pyramids at various scales. Their experiments on Cross-eyed database achieved a GMR of $100$\% at an FMR of $0.01$\%.

Sushree \textit{et al.} \cite{Sushree-2017} have proposed an approach to match NIR and VIS periocular images using difference of Gaussian (DoG) filtering based illumination normalization. HOG feature vectors are then extracted from the normalized images and matched using Cosine similarity metrics. Their method achieves GARs of $25.03$\% and $83.12$\% at an FAR value of $0.1$ on IMP and PolyU database respectively using both left and right periocular images. The authors have also compared their results with that of the well known LBP features and found out that the HOG features outperforms the LBP operator for both databases. 

As can be seen from the above discussion, the existing works have mostly focused on general machine learning approaches involving feature extraction followed by either matching or classifying these features using a classifier. The performance on IMP database is very poor due to presence of poor quality images in it, whereas experiments on Cross-eyed database produce very high accuracy values as the images present in it are of very high quality. However, it can not be assured that, algorithms, which achieve higher performance accuracies on this database can be implemented in practical scenarios. PolyU cross-spectral iris databases has achieved a maximum GAR value of $83.12$\% at an FAR value of $0.1$. Further research in this direction should focus on increasing the recognition accuracies of this database using advanced methods as well as on creating more such databases pertaining to practical surveillance applications. 
\section{Cross-spectral Periocular Databases}
There are a large number of methodologies available for periocular biometrics, but very few number of databases are being designed for this purpose. Most researchers have used face and iris databases, since periocular region is present in such databases. The face databases used in periocular research mostly include images in VIS range \cite{Park-2011}, \cite{Santos-2013} where the face regions are then cropped in order to get the periocular images. However, the iris databases used for this purpose include NIR images \cite{Fernandez-2015} where raw iris images are used. Some heterogeneous face databases containing images in VIS, NIR and SWIR spectra have also been used in literature \cite{cao:1}, \cite{cao:2} and \cite{cao:3}, but, these databases are not publicly available. Recently, some databases for cross-spectral periocular region have been made public. This section describes these databases briefly, protocols used, their advantages and disadvantages. A summary of them is presented in Table \ref{overview}. It not only summarizes the publicly available databases used in cross-spectral research but also lists the best accuracies obtained on each of them. 


\begin{table}[h]
\centering
\caption{Overview of Existing Cross-spectral Periocular Databases}
\label{overview}
\begin{tabular}{lllllll}
\hline
Name       & \begin{tabular}[c]{@{}l@{}}No. of\\ subjects\end{tabular} & Illumination                                                     & \begin{tabular}[c]{@{}l@{}}No. of images\\ per subject\end{tabular} & Size of images                                                               & \begin{tabular}[c]{@{}l@{}}Total no. of\\ Images\end{tabular} & \begin{tabular}[c]{@{}l@{}}Best reported \\ performance\end{tabular}  \\ \hline
\begin{tabular}[c]{@{}l@{}}IMP \cite{Sharma-2014}\\ (2014)\end{tabular}& 62                                                        & \begin{tabular}[c]{@{}l@{}}VIS\\ NIR\\ Night vision\end{tabular} & \begin{tabular}[c]{@{}l@{}}5\\ 10\\ 5\end{tabular}                  & \begin{tabular}[c]{@{}l@{}}301-by-601\\ 640-by-480\\ 540-by-640\end{tabular} & 1240                                                          & \begin{tabular}[c]{@{}l@{}}GAR of 47.08 \% \\ at 1\% FAR\end{tabular} \\
\begin{tabular}[c]{@{}l@{}}PolyU \cite{Ramaiah-2016}\\ (2016)\end{tabular}     & 209                                                       & \begin{tabular}[c]{@{}l@{}}VIS\\ NIR\end{tabular}                & \begin{tabular}[c]{@{}l@{}}15\\ 15\end{tabular}                     & \begin{tabular}[c]{@{}l@{}}480-by-640\\ 480-by-640\end{tabular}              & 6270                                                          & \begin{tabular}[c]{@{}l@{}}GAR of 83.12\%\\ at 10\% FAR\end{tabular}  \\
\begin{tabular}[c]{@{}l@{}}Cross-eyed \cite{Crosseyed:1}\\ (2016)\end{tabular} & 120                                                       & \begin{tabular}[c]{@{}l@{}}VIS\\ NIR\end{tabular}                & \begin{tabular}[c]{@{}l@{}}8\\ 8\end{tabular}                       & \begin{tabular}[c]{@{}l@{}}800-by-900\\ 800-by-900\end{tabular}              & 3840                                                          & GF2 of 0.74\%                                                         \\ \hline
\end{tabular}
\end{table}
\subsection{IMP Database}
\begin{figure}[h!]
\centering
\includegraphics[width=0.8\linewidth]{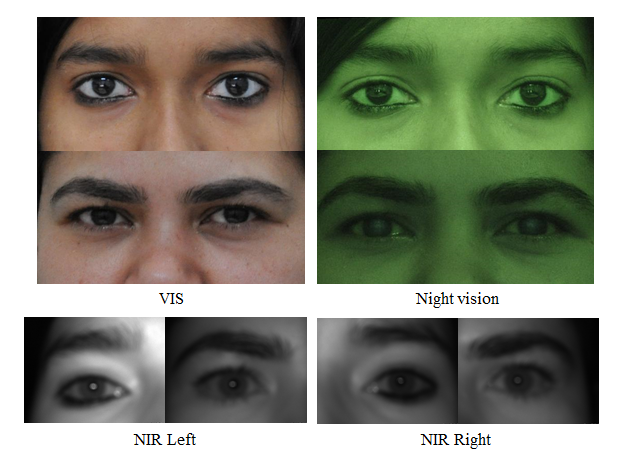}
\caption{Sample periocular images from IMP database \cite{Sharma-2014}.}
\label{fig:imp_images}
\end{figure}

The IMP \cite{Sharma-2014}  database contains periocular images captured in VIS, NIR and night vision spectra. A total of 62 subjects, with 5 images captured in each spectrum are present. The VIS and night vision images comprise both left and right periocular regions whereas the NIR images contain separate images for left and right periocular regions. Fig. \ref{fig:imp_images} shows sample VIS and NIR periocular images of two different subjects in this database. The VIS and night vision images have dimensions of $301\times601$ pixels. On the other hand, each of the NIR images has a dimension of $640\times480$ pixels. Therefore, a total of 1240 images are present with 310, 310, and 620 images from VIS, night vision and NIR spectra respectively.
The IMP dataset is claimed to be the first multi-spectral periocular dataset which was made public in the year 2014. As can be seen from the figure, the periocular images in NIR illumination are of very poor resolution which causes degradation of the recognition performance of the VIS-NIR matching. For this reason, the highest accuracy achieved till now is a GAR of $47.08$\% at $1$\% FAR.  

\subsection{PolyU Cross-spectral Iris Database}
\begin{figure}[h!]
\centering
\includegraphics[width=0.8\linewidth]{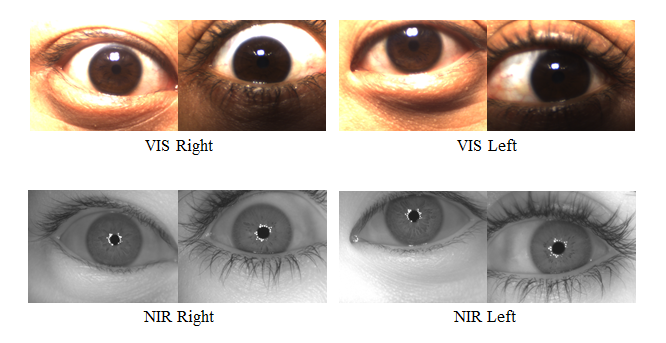}
\caption{Sample periocular images from PolyU cross-spectra iris database \cite{Ramaiah-2016}.}
\label{fig:poly_images}
\end{figure}

The PolyU cross-spectral iris database \cite{Ramaiah-2017} consists of left and right periocular images from 209 subjects in VIS and NIR spectra. This database contains 15 images per spectrum per subject thereby making the total number of images to be 6270. Each of these images has a size of $640\times480$ pixels. Fig. \ref{fig:poly_images} shows the sample periocular images from PolyU database.
This database contains the highest number of subjects making it the largest periocular database existing in the literature in heterogeneous domain. The appearance of VIS and NIR images largely vary in terms of illumination and alignment.

\subsection{Crosseyed Periocular Database}
\begin{figure}[h!]
\centering
\includegraphics[width=0.8\linewidth]{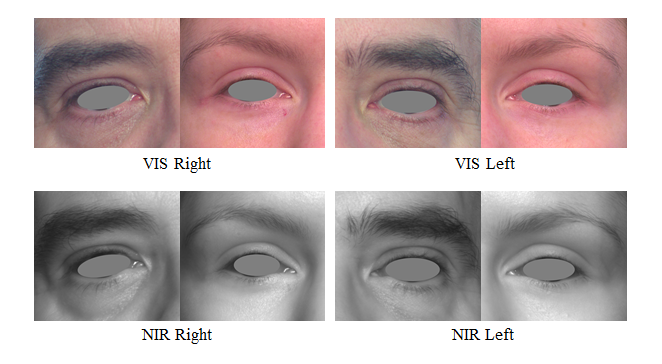}
\caption{Sample periocular images from Cross-eyed periocular database \cite{Crosseyed:1}.}
\label{fig:cross_images}
\end{figure}

Like PolyU cross-spectral iris database, the Cross-eyed periocular database \cite{Crosseyed:1} contains images of both left and right periocular regions in both VIS and NIR spectra. A total of 3840 images from 120 subjects with 16 images (8 images from each region) per subject per spectrum are present in this database. Fig. \ref{fig:cross_images} shows the sample images of a subject for left and right periocular regions in VIS and NIR spectrum. The dimension of each of these images is of $800\times900$ pixels.
As can be seen from the figure, the sclera regions are masked in order to avoid the use of iris information while performing experiments for the periocular recognition. Although acquired separately, left and right periocular images are perfectly aligned. Due to these reasons, the recognition accuracies obtained in this database is very high, however such a database is not useful while considering practical scenarios. 

\section{Conclusion and Future Work}

Practical surveillance applications require the acquisition of images in a wavelength other than VIS light. However, installation of systems in order to recognize individuals completely in those domains is nearly impossible as it is both time and cost consuming. Hence, algorithms are required in order to match images from different domain. Not only face and iris, but periocular recognition has also been explored in the direction of cross-spectral matching due to its advantages over face and iris biometrics. In this paper, a detailed overview of existing works on cross-spectral periocular recognition has been given. In addition to this, some insights are given into existing periocular databases on cross-spectral domain which include images in VIS, NIR, and night vision spectra. However, the number of databases available publicly is very limited with very poor image quality. A number of works have been done on this topic, focusing mostly on extracting local and global feature descriptors from the pre-processed periocular images followed by matching them using some distance metrics. The pre-processing blocks are mostly designed to alleviate the appearance differences between gallery and probe images. Research works have also taken into consideration practical challenges such as long standoff distances and varying ranges of spectral bands. Still the recognition accuracy of heterogeneous biometrics is some step behind, especially for the case of IMP database \cite{Sharma-2014}.

This work is expected to provide insights into future research directions in this area. To mention a few, researchers can explore advanced algorithms on machine learning such as deep learning to improve the state-of-the-art accuracies on existing challenging databases. Special attention can be given on developing algorithms that can improve the performance of images with poor resolution and improper alignment between images of different domain. Multimodal biometric approach can also be utilized in order to increase the recognition accuracy of individuals. In addition to that, more databases consisting of periocular images in heterogeneous spectral bands can be developed with added challenges such as varying standoff distances as well as varying spectral ranges. 

%
%


\begin{thebibliography}{6}
%

\bibitem{Park-2009}
Park, U., Ross, A. and Jain, A. K.: Periocular biometrics in the visible spectrum: A feasibility study. In: 3rd IEEE International Conference on Biometrics: Theory, Applications, and Systems (BTAS), pp. 1–-6 (2009). 

\bibitem{Mandal2}
B. Mandal and X. D. Jiang and A. Kot: Verification of Human Faces Using Predicted Eigenvalues, $19^{th}$ International Conference on Pattern Recognition (ICPR), Tempa, Florida, USA, Dec, 2008, pages 1-4 (2008).

\bibitem{Mandal3}
B. Mandal and Liyuan Li and V. Chandrasekhar and Joo Hwee Lim: Whole Space Subclass Discriminant Analysis for Face Recognition, International Conference on Image Processing (ICIP), Quebec city, Canada, Sep, 2015, pages 329-333 (2015).

\bibitem{Mandal4}
B. Mandal, Shue{-}Ching Chia, Liyuan Li, Vijay Chandrasekhar, Cheston Tan and Joo Hwee Lim: A Wearable Face Recognition System on Google Glass for Assisting Social Interactions, $12^{th}$ Asian Conference on Computer Vision (ACCV) Workshops, Singapore, Nov, 2014, pages 419--433 (2014).


\bibitem{mandal2016trends}
Mandal, Bappaditya, Lim, Rosary Yuting, Dai, Peilun, Sayed, Mona Ragab, Li, Liyuan, Lim and Joo Hwee: Trends in Machine and Human Face Recognition. In Advances in Face Detection and Facial Image Analysis, pp. 145--187, Springer International Publishing (2016).

\bibitem{url:1}
Greenberg, A., Greenberg, A., Barrett, B., Barber, G., Graff, G., Barrett, B., Newman, L. and Lapowsky, I. Watch a 10-Year-Old's Face Unlock His Mom's iPhone X. [online] WIRED. Available at: https://www.wired.com/story/10-year-old-face-id-unlocks-mothers-iphone-x/ (14/11/2017).

\bibitem{Mandal5}
B. Mandal: Face recognition: Perspectives from the real world, $14^{th}$ International Conference on Control, Automation, Robotics and Vision (ICARCV), Phuket, Thailand, Nov, 2016, pages 1--5 (2016).


\bibitem{Bharadwaj-2010}
Bharadwaj, Bhatt S., H. S., Vatsa, M. and Singh, R.,: Periocular biometrics: When iris recognition fails. In: 4th IEEE International Conference on Biometrics: Theory Applications and Systems (BTAS), pp. 1--6 (2010). 

\bibitem{Alonso-2016}
Fernando, Alonso-Fernandez and Bigun, Josef: A survey on periocular biometrics research. Pattern Recognition Letters, vol. 82, pp. 92--105, Elsevier (2016).

\bibitem{Nigam-2015}
Nigam, I., Vatsa, M. and Singh, R.: Ocular biometrics: A survey of modalities and fusion approaches. Information Fusion, vol. 26, pp. 1–35 (2015).

\bibitem{Hyunju-2011}
Maeng, Hyunju, Choi, Hyun-Cheol, Park, Unsang, Lee, Seong-Whan and Jain, A. K.,: NFRAD: Near-Infrared Face Recognition at a Distance. In: International Joint Conference on Biometrics (IJCB), pp. 1--7 (2011). 

\bibitem{zhu-2014}
Zhu, Jun-Yong, Zheng, Wei-Shi, Lai, Jian-Huang and Li, Stan Z,: Matching NIR face to VIS face using transduction. IEEE Transactions on Information Forensics and Security, \textbf{9} (3), pp. 501--514 (2014).

\bibitem{Juefei-2015}
Juefei-Xu, Felix, Pal, Dipan K. and Savvides, Marios,: NIR-VIS Heterogeneous Face Recognition via Cross-Spectral Joint Dictionary Learning and Reconstruction. In: IEEE Conference on Computer Vision and Pattern Recognition (CVPR) Workshops (2015).

\bibitem{Ross-2009}
Ross, A., Pasula, R. and Honark, L.,: Exploring multispectral iris recognition beyond 900nm. In: 3rd IEEE International Conference on Biometrics: Theory, Applications, and Systems (BTAS), pp. 1--8 (2009). 

\bibitem{Zuo-2010}
Zuo, J.. Nicolo, F. and Schmid, N. A.,: Cross spectral iris matching based on predictive image mapping. In: 4th IEEE International Conference on Biometrics: Theory, Applications, and Systems (BTAS), pp. 1--5 (2010). 

\bibitem{Ramaiah-2017}
Ramaiah, N. and Kumar, A.: Towards more accurate iris recognition using cross-spectral matching. IEEE Transactions on Image Processing, \textbf{26} (1), pp. 208–-221 (2017).

\bibitem{Sharma-2014}
Sharma, A., Verma, S., Vasta, M. and Singh, R.: On cross spectral periocular recognition. In: IEEE International Conference on Image Processing (ICIP), pp. 5007--5011 (2014). 

\bibitem{cao:1}
Cao, Z. X. and Schmid, N. A.: Matching heterogeneous periocular regions: Short and long standoff distances. In: IEEE International Conference on Image Processing (ICIP), pp. 4967--4971 (2014). 

\bibitem{cao:2}
Cao, Z and Schmid, Natalia A.: Recognition performance of cross-spectral periocular biometrics and partial face at short and long standoff distance. Open Transactions on Information Processing, \textbf{1} (2), pp. 20--32 (2014).

\bibitem{cao:3}
Cao, Zhicheng and Schmid, Natalia A: Fusion of operators for heterogeneous periocular recognition at varying ranges. Pattern Recognition Letters, vol. 82, pp. 170–-180, Elsevier (2016).

\bibitem{Ramaiah-2016}
Ramaiah, N. P. and Kumar, A.: On matching cross-spectral periocular images for accurate biometric identification. In: 8th IEEE International Conference on Biometrics: Theory, Applications, and Systems (BTAS) (2016).

\bibitem{Raja-2016}
Raja, K. B., Raghavendra, R. and Busch, C.: Cross-spectrum periocular authentication for NIR and visible images using bank of statistical filters. In: IEEE International Conference on Imaging Systems and Techniques (IST), pp. 227--231 (2016). 

\bibitem{Raja-2017}
Raja, K and Raghavendra, R. and Busch, C.: Scale-level score fusion of steered pyramid features for cross-spectral periocular verification. In: $20^{th}$ International Conference on Information Fusion (Fusion), pp. 1--7 (2017). 

\bibitem{Crosseyed:1}
Sequeira, A., Chen, L., Wild, P., Ferryman. J., Alonso-Fernandez, F., Raja, K. B., Raghavendra, R., Busch, C. and Bigun, J.: Cross-Eyed - Cross-Spectral Iris/Periocular Recognition Database and Competition. In: International Conference of the Biometrics Special Interest Group (BIOSIG), pp.1–-5 (2016). 

\bibitem{Crosseyed:2}
Sequeira, A. F., Chen, L., Ferryman, J., Wild, P., Alonso-Fernandez, F., Bigun, J., Raja, K. B., Raghavendra, R., Busch, C. Pereira, T. de Freitas, Marcel, S., Behera, S. S., Gour, M. and Kanhangad, V.: Cross-eyed 2017: Cross-spectral iris/periocular recognition competition. In: IEEE International Joint Conference on Biometrics (IJCB), pp. 725–-732 (2017). 

\bibitem{Sushree-2017}
Behera, S. S., Gour, M., Kanhangad, V. and Puhan, N.: Periocular recognition in cross-spectral scenario. In: IEEE International Joint Conference on Biometrics (IJCB), pp. 681–-687 (2017). 

\bibitem{Park-2011}
Park, U., Jillela, R. R., Ross, A. and Jain, A. K.: Periocular Biometrics in the Visible Spectrum. IEEE Transactions on Information Forensics and Security, \textbf{6} (1), pp. 96--106 (2011).

\bibitem{Santos-2013}
Santos, G. and Proença, H.,: Periocular biometrics: An emerging technology for unconstrained scenarios. In: IEEE Workshop on Computational Intelligence in Biometrics and Identity Management (CIBIM), pp. 14--21 (2013).

\bibitem{Fernandez-2015}
Alonso-Fernandez, Fernando and Bigun, Josef,: Near-infrared and visible-light periocular recognition with gabor features using frequency-adaptive automatic eye detection. IET Biometrics, \textbf{4} (2), pp. 74--89 (2015).



\end{thebibliography}
\end{document}